\documentclass[11pt,letter]{article}
\usepackage{times}
\usepackage{amsmath, amsthm, amssymb}
\usepackage[latin1]{inputenc}
\usepackage[T1]{fontenc}
\usepackage[english]{babel}

\usepackage{graphicx, color}
\usepackage{xspace, url}
\usepackage{setspace}

\newcommand{\kinships}{Kinships\xspace}
\newcommand{\umls}{UMLS\xspace}
\newcommand{\nations}{Nations\xspace}

\newcommand{\s}{{\cal E}}
\newcommand{\lhs}{$lhs$\xspace}
\newcommand{\rel}{$rel$\xspace}
\newcommand{\rhs}{$rhs$\xspace}

\newcommand{\sme}{{\sf SME}\xspace}
\newcommand{\smel}{{\sf SME}(linear)\xspace}
\newcommand{\smeb}{{\sf SME}(bilinear)\xspace}

\newcommand{\lfm}{{\sf LFM}\xspace}

\newcommand{\irm}{{\sf IRM}\xspace}

\newcommand{\rescal}{{\sf RESCAL}\xspace}
\newcommand{\cp}{{\sf CP}\xspace}

\newcommand{\mrc}{{\sf MRC}\xspace}

\begin{document}

\title{\vspace*{-2cm}A Semantic Matching Energy Function for Learning\\
\vspace*{-0.2cm}
 with Multi-relational Data}
\author{\vspace*{-0.1cm}Xavier Glorot$^{(1)}$, Antoine Bordes$^{(2)}$, Jason Weston$^{(3)}$, Yoshua Bengio$^{(1)}$\\
\vspace*{-0.2cm}
\hspace{-0.6cm}
\begin{tabular}{l}
\vspace*{-0.2cm}
{\small $^{(1)}$} {\small DIRO, Universit\'e de Montr\'eal, Montr\'eal, QC, Canada} -- \small{\texttt{\{glorotxa,bengioy\}@iro.umontreal.ca}}\\
\vspace*{-0.2cm}
{\small $^{(2)}$} {\small CNRS - Heudiasyc, Universit\'e de Technologie de Compi\`egne, France} -- \small{\texttt{bordesan@hds.utc.fr}}\\
{\small $^{(3)}$} {\small Google, New York, NY, USA} -- \small{\texttt{jweston@google.com}}\\
\end{tabular}
}
\date{}

\maketitle
\thispagestyle{empty}
\pagestyle{empty}

\section{Introduction}
\vspace*{-0.3cm}

Multi-relational data, which refers to graphs whose nodes represent entities and edges correspond
to relations that link these entities, plays a pivotal role in many areas such as
recommender systems, the Semantic Web, or computational biology. 
Relations are modeled as triplets of the form (subject, relation, object), where a
relation either models the relationship between two entities or between an entity and an
attribute value; relations are thus of several types.
%
%
%
%
In spite of their appealing ability for representing complex data, multi-relational graphs
remain complicated to manipulate for several reasons (noise, heterogeneity, large-scale dimensions, etc.), and
%
conveniently represent, summarize or de-noise this kind of data is now a central 
challenge in statistical relational learning~\cite{Getoor:2007}.

In this work, we propose a new model to learn multi-relational semantics, that is, to
encode multi-relational graphs into representations that capture
the inherent complexity in the data, while seamlessly defining similarities among entities and relations 
and providing predictive power.
Our work is based on an original energy function, 
 which is trained to assign low energies to plausible triplets of a
multi-relational graph.
This energy function, termed {\it semantic matching energy}, relies on a compact distributed
representation: all elements (entity and relation type) are represented into the same relatively low (e.g. 50)
dimensional embedding vector space. 
The embeddings are learnt by a neural network whose particular architecture 
and training process force them to capture the structure implicit in the training data
and generalize the graph formed from training triplets.
Unlike in previous work \cite{Kemp:2006,Sutskever:2009,Nickel:2011,Jenatton:2012}, in this model, relation types are modeled similarly as entities. In this way, entities can also play the role of relation type, as in natural language for instance, and this requires less parameters when the number of relation types grows.
%
%
We show empirically that this model achieves competitive results on benchmark tasks of link prediction, i.e., generalizing
outside of the set of given valid triplets.

\vspace*{-0.3cm}
\section{Semantic Matching Energy Function} \label{sec:model}
\vspace*{-0.3cm}



This work considers multi-relational databases as graph models.
%
To each individual
node of the graph corresponds an element of the database, which we term an {\em entity},
and each link defines a {\em relation} between entities.  Relations are directed and there
are typically several different kinds of relations.
Let ${\cal C}$ denote the dictionary which includes all entities and relation types, and
let ${\cal R} \subset {\cal C}$ be the subset of entities which are relation types.
A relation is denoted by a triplet (\lhs, \rel, \rhs),
where \lhs is the {\em left} entity, \rhs the {\em right} one and \rel the {\em type} of
relation between them.

\vspace*{-0.2cm}
\subsection{Main ideas}
\vspace*{-0.3cm}
The main ideas behind our semantic matching energy function are the
following.

\begin{itemize}
\vspace*{-0.2cm}
\item Named symbolic entities (entities {\it and} relation types) are associated with a
  $d$-dimensional vector space, termed the ``embedding space''.  The $i^{th}$
  entity is assigned a vector $E_i \in {\mathbb R}^d$. Note that more general
  mappings from an entity to its embedding are possible.
  %
\item The semantic matching energy value associated with a particular triplet (\lhs, \rel,
  \rhs) is computed by a parametrized function $\s$ that starts by mapping all symbols to
  their embeddings and then combines them in a structured fashion. Our model is termed
  ``semantic matching'' because $\s$ relies on a
  matching criterion computed between both sides of the triplet. 
\item The energy function $\s$ is optimized to be lower for training examples than for
  other possible configurations of symbols.  

\end{itemize}


\vspace*{-0.2cm}
\subsection{Neural network parametrization}\label{sec:modelnn}
\vspace*{-0.3cm}


The energy function $\s$ (denoted \sme) is encoded using a neural network, whose architecture first processes each entity in parallel, like
in siamese networks \cite{Bromley-bentz-93}. The intuition is that
the relation type should first be used to extract relevant components
from each argument's embedding, and put them in a space where they can
then be compared.

\begin{itemize}
\item[(1)] Each symbol of the input triplet (\lhs, \rel, \rhs) is mapped to its embedding
  $E_{lhs}$, $E_{rel}$, $E_{rhs}$ $\in {\mathbb R}^d$.
\item[(2)] The embeddings $E_{lhs}$ and $E_{rel}$ respectively associated with the $lhs$
  and $rel$ arguments are used to construct a new relation-dependent embedding
  $E_{lhs(rel)}$ for the $lhs$ in the context of the relation type represented by
  $E_{rel}$, and similarly for the $rhs$: $E_{lhs(rel)} = g_{left}(E_{lhs},E_{rel})$ and
  $E_{rhs(rel)} = g_{right}(E_{rhs},E_{rel})$, where $g_{left}$ and $g_{right}$ are
  parametrized functions whose parameters are tuned during training. The dimension of $E_{lhs(rel)}$ and $E_{rhs(rel)}$,
  which we denote $p$, is low-dimensional but not necessarily equal to $d$, the dimension of the entity embedding space.
\item[(3)] The energy is computed by "matching" the transformed embeddings of the
  left-hand and right-hand sides: $\s((lhs,rel,rhs)) = h(E_{lhs(rel)}, E_{rhs(rel)})$,
  $h$ is a dot product in our experiments.
\end{itemize}


%
We studied two options for the $g$ functions, which lead to two
versions of \sme:
\begin{itemize}

\item {\it Linear form} (denoted \smel), in this case $g$ functions are
  simply linear layers:
  {\small
  \begin{eqnarray*}
    E_{lhs(rel)} &= g_{left}(E_{lhs},E_{rel}) &= W_{l1}E_{lhs}^\intercal + W_{l2}E_{rel}^\intercal + b_l^\intercal.\\
    E_{rhs(rel)} &= g_{right}(E_{rhs},E_{rel}) &= W_{r1}E_{rhs}^\intercal + W_{r2}E_{rel}^\intercal +b_r^\intercal.
  \end{eqnarray*}}
  with $W_{l1}$, $W_{l2}$, $W_{r1}$, $W_{r2}$ $\in {\mathbb R}^{p\times d}$,
  $b_l$, $b_r$ $\in {\mathbb R}^{p}$ and $E^\intercal$ denotes the transpose of
  $E$.
  This leads to the energy:
  {\small
  $
    \s((lhs,rel,rhs)) = - \left(W_{l1}E_{lhs}^\intercal + W_{l2}E_{rel}^\intercal+b_l^\intercal \right)^\intercal \left(W_{r1}E_{rhs}^\intercal + W_{r2}E_{rel}^\intercal + b_r^\intercal \right) $.
  }

\item {\it Bilinear form} (denoted \smeb), $g$ functions are
  using 3-modes tensors as core weights:
 {\small
  \begin{eqnarray*}
    E_{lhs(rel)} &= g_{left}(E_{lhs},E_{rel}) &= \left(W_{l}\bar{\times}_3 E_{rel}^\intercal\right)E_{lhs}^\intercal +b_l^\intercal.\\
    E_{rhs(rel)} &= g_{right}(E_{rhs},E_{rel}) &= \left(W_{r}\bar{\times}_3 E_{rel}^\intercal\right)E_{rhs}^\intercal +b_r^\intercal.     
  \end{eqnarray*}
  }
  with $W_{l}$, $W_{r}$ $\in {\mathbb R}^{p\times d \times d}$ (weights) and $b_l$, $b_r$
  $\in {\mathbb R}^{p}$ (biases). $\bar{\times}_3$ denotes the $n$-mode vector-tensor
  product along the $3^{rd}$ mode.
  This leads to the following form for the energy:
  {\small
    $\s((lhs,rel,rhs)) = - \left( \left(W_{l}\bar{\times}_3 E_{rel}^\intercal \right)E_{lhs}^\intercal +b_l^\intercal\right)^\intercal \left(\left( W_{r}\bar{\times}_3 E_{rel}^\intercal\right)E_{rhs}^\intercal +b_r^\intercal\right) $.
  }
\end{itemize}

To train the parameters of the energy function $\s$ we loop over all of the training data
resources and use stochastic gradient descent with a ranking objective inspired by~\cite{wsabie}.

\vspace*{-0.3cm}
\section{Empirical Evaluation}\label{sec:exp}
\vspace*{-0.3cm}


\begin{table}[t]
\vspace*{-0.3cm}
  \begin{center}
      \caption{\label{tab:data_bench}  Statistics of
        datasets used in this paper.
      }
    {\small    
    \begin{tabular}{|l|c|c|c|c|}
      \hline
      Dataset   & \bf Nb. of relation & \bf Nb. of  & \bf Nb. of observed & \bf \% valid relations \\ 
      & \bf types & \bf entities & \bf relations  & \bf in obs. ones\\
      \hline
      \hline
      \bf \umls     & 49 & 135 & 893,025 & 0.76\\
      \bf \kinships & 26 & 104 & 281,216 & 3.84 \\
      \bf \nations & 56 & 14 & 11,191 & 22.9 \\
      \hline
    \end{tabular}
    }
  \end{center}
  \vspace*{-0.3cm}
\end{table}

To evaluate against existing methods, we performed link prediction experiments on benchmarks
from the literature, whose statistics are in Table~\ref{tab:data_bench}.

The link prediction task consists in predicting whether two entities should be connected by a given
relation type. This is useful for completing missing values of a graph, forecasting the
behavior of a network, etc. but also to assess the quality of a representation.
We evaluate our model on \umls, \nations and \kinships, following the setting introduced
in \cite{Kemp:2006}.
%
The standard evaluation metric is area under the precision-recall curve (AUC).
%
%
Table~\ref{fig:lpred} presents results of \sme along with those of \rescal, \mrc,
\irm, \cp (CANDECOMP-PARAFAC) and \lfm, which have been extracted from 
\cite{Nickel:2011,Jenatton:2012}.
%

%
The linear formulation of \sme is outperformed by \smeb on all three tasks.
The largest differences for \nations and \kinships indicate that, for these problems, 
a joint interaction between both \lhs, \rel and \rhs is crucial to represent the data well: 
relations cannot be simply decomposed as a sum of bigrams.
This is particularly true for the complex kinship systems of the Alyawarra. 
On the contrary, interactions within the \umls network can be represented by simply 
considering the various (entity,entity) and (entity,relation type) bigrams.
Compared to other methods, \smeb performs similarly to \lfm on \umls but is slightly outperfomed on \nations.
%
%
On \kinships, it is outperformed by \cp, \rescal and \lfm: on this dataset with complex ternary interactions, either the training process
of the tensor factorization methods, based on reconstruction, or the combination of bigram and trigram interactions seems to be beneficial compared to our predictive approach.
%
%
%
Compared to \mrc, which is not using a matrix-based encoding, \smeb is highly competitive.
%


\begin{table}[h!]
  \vspace*{-0.3cm}
\newcommand{\mc}[3]{\multicolumn{#1}{#2}{#3}}
  \begin{center}
  \begin{small}
      \caption{\label{fig:lpred} Comparisons of area under the precision-recall curve 
    (AUC) for link prediction.}
    \begin{tabular}{|l|l@{\:}l|l@{\:}l|l@{\:}l|}
    \hline
    Method &  \mc{2}{c|}{\bf UMLS} & \mc{2}{c|}{\bf Nations} & \mc{2}{c|}{\bf Kinships}\\
    \hline
    \hline
    \smel & {\bf 0.983} &$\pm$ 0.004 & 0.777 &$\pm$ 0.025 & 0.149 &$\pm$ 0.003\\
    \smeb & {\bf 0.985} &$\pm$ 0.003 & 0.865 &$\pm$ 0.015 & 0.894 &$\pm$ 0.011 \\
    \lfm & {\bf 0.990} &$\pm$ 0.003 & {\bf 0.909} &$\pm$ 0.009 & \bf 0.946 &$\pm$ 0.005 \\
    \rescal &  0.98 && 0.84 && \bf 0.95 &\\
    \cp & 0.95 && 0.83 && \bf 0.94 &\\
    \mrc &  0.98 && 0.75 && 0.85 &\\
    \irm & 0.70 && 0.75 && 0.66 &\\
    \hline
    \end{tabular}
  \end{small}
  \vspace*{-0.6cm}
  \end{center}
\end{table}

Even if experimental results on these benchmarks are mixed, it is worth noting that, contrary to all previous methods, SME models relation types as vectors, lying in the same space as entities. From a conceptual viewpoint, this is powerful, since it models any relation types as a standard entity (and vice-versa). Hence, SME is the only method that could be directly applied on data for which any entity can also create relationships between other entities.

\subsection*{Acknowledgements}
\vspace*{-0.3cm}
This work was supported by the French ANR (EVEREST-12-JS02-005-01), the Pascal2 European NoE, the DARPA DL Program, NSERC, CIFAR, the Canada Research Chairs, and Compute Canada.

{{\small
\bibliographystyle{plain}
\bibliography{semantic_parsing,aaai_paper,ebrm_mlj}}}

\end{document}